\documentclass{article}

\usepackage[nonatbib,final]{nips_2017} 

\usepackage{graphicx} 
\usepackage{subfigure} 

\usepackage{algorithm}
\usepackage{algorithmic}

\usepackage{amsmath,amsfonts,amssymb}

\newtheorem{theorem}{Theorem}

\usepackage{booktabs}
\usepackage{footnote}
\makeatletter
\def\@xfootnote[#1]{%
  \protected@xdef\@thefnmark{#1}%
  \@footnotemark\@footnotetext}
\makeatother

\usepackage{hyperref}

\title{ProjectionNet: Learning Efficient On-Device Deep Networks Using Neural Projections}

\author{
  Sujith Ravi\\
  Google Research, Mountain View, CA, USA \\
  \texttt{sravi@google.com}
}

\begin{document} 

\maketitle

\begin{abstract}
Deep neural networks have become ubiquitous for applications related to visual recognition and language understanding tasks. However, it is often prohibitive to use typical neural network models on devices like mobile phones or smart watches since the model sizes are huge and cannot fit in the limited memory available on such devices. While these devices could make use of machine learning models running on high-performance data centers with CPUs or GPUs, this is not feasible for many applications because data can be privacy sensitive and inference needs to be performed directly ``on'' device.

We introduce a new architecture for training compact neural networks using a joint optimization framework. At its core lies a novel objective that jointly trains using two different types of networks--a full trainer neural network (using existing architectures like Feed-forward NNs or LSTM RNNs) combined with a simpler ``{\it projection}'' network that leverages random projections to transform inputs or intermediate representations into bits. The simpler network encodes lightweight and efficient-to-compute operations in bit space with a low memory footprint. The two networks are trained jointly using backpropagation, where the projection network learns from the full network similar to apprenticeship learning. Once trained, the smaller network can be used directly for inference at low memory and computation cost.
We demonstrate the effectiveness of the new approach at significantly shrinking the memory requirements of different types of neural networks while preserving good accuracy for several visual recognition and text classification tasks. We also study the question ``how many neural bits are required to solve a given task?'' using the new framework and show empirical results contrasting model predictive capacity (in bits) versus accuracy on several datasets. Finally, we show how the approach can be extended to other learning settings and derive projection models optimized using graph structured loss functions.
\end{abstract}

\section{Introduction}
\label{sec:intro}
Recent advances in deep neural networks have resulted in powerful models that demonstrate high predictive capabilities on a wide variety of tasks from image classification~\cite{krizhevsky2012imagenet} to speech recognition~\cite{hinton2012deep} to sequence-to-sequence learning~\cite{sutskever2014sequence} for natural language applications like language translation~\cite{nmt2016bahdanau}, semantic conversational understanding~\cite{smartreply2016} and other tasks. These networks are typically large, comprising multiple layers involving many parameters, and trained on large amounts of data to learn useful representations that can be used to predict outputs at inference time. For efficiency reasons, training these networks is often performed with high-performance distributed computing involving several CPU cores or graphics processing units (GPUs).

In a similar vein, applications running on devices such as mobile phones, smart watches and other IoT devices are on the rise. Increasingly, machine learning models are used to perform real-time inference directly on these devices---e.g., speech recognition on mobile phones~\cite{Schuster2010}, medical devices to provide real-time diagnoses~\cite{medical} and Smart Reply on watches~\cite{smartwear}, among others. However, unlike high-performance clusters running on the cloud, these devices operate at low-power consumption modes and have significant memory limitations. As a result, running state-of-the-art deep learning models for inference on these devices can be very challenging and often prohibitive due to the high computation cost and large model size requirements that exceed device memory capacity. Delegating the computation-intensive operations from device to the cloud is not a feasible strategy in many real-world scenarios due to connectivity issues (data cannot be sent to the server) or privacy reasons (certain types of data and processing needs to be restricted to a user's personal device). In such scenarios, one solution is to take an existing trained neural network model and then apply compression techniques like quantization (e.g., reducing floating point precision~\cite{CourbariauxBD14}) to reduce model size. However, while these techniques are useful in some cases, applying them post-learning to a complex neural network tends to dilute the network's predictive quality and does not yield sufficiently high performance. An alternate strategy is to train small models for on-device prediction tasks, but these can lead to significant drop in accuracies~\cite{Chun2009} which limits the usability of such models for practical applications. In particular, feature or vocabulary pruning techniques commonly applied to limit parameters in models like recurrent neural networks, while yielding lower memory footprint, can affect the predicitive capacity of the network for language applications.

This motivates the learning of efficient on-device machine learning models with low memory footprint that can be run directly on device for inference at low computation cost.
\paragraph{Contributions.} In this paper, we propose a joint learning framework based on neural projections to learn lightweight neural network models for performing efficient inference on device.

\begin{itemize}
\item The neural projection framework can leverage any existing deep network like feed-forward or recursive neural network to teach a lightweight projected model in a joint optimization setup which is trained end-to-end using backpropagation. We use projections based on locality sensitive hashing to represent the hidden units for the lightweight network which encodes operations that are extremely efficient to compute during inference (Section~\ref{sec:randnn}).

\item The framework permits efficient distributed training but is optimized to produce a neural network model with low memory footprint that can run on devices at low computation cost. The model size is parameterized and configurable based on the task or device capacity.
\item We demonstrate the effectiveness of the new approach in achieving significant reduction in model sizes while providing competitive performance on multiple visual and language classification tasks (Section~\ref{sec:exp_rnn}).
\item The proposed framework is further used to study and characterize the predictive capacity of existing deep networks in terms of the number of {\it neural projection bits} required to represent them compactly (Section~\ref{sec:neural_bits}).
\end{itemize}

\section{Related Work}
\label{sec:related}
There have been a number of related works in the literature that attempt to learn efficient models under limited size or memory constraints. Some of these works employ techniques ranging from simple dictionary lookups to feature pruning~\cite{stolcke2000} or hashing~\cite{Weinberger2009,Shi2009,ganchev2008} to neural network compression. In the past, researchers have proposed different methods to achieve compact representations for neural networks using reduced numerical precision~\cite{CourbariauxBD14}, vector quantization~\cite{GongLYB14}, binarization strategies for networks~\cite{binarynet} or weight sharing~\cite{Denil2013,hashednets}. Most of these methods aim to exploit redundancy in the network weights by grouping connections using low-rank decomposition or hashing tricks. In contrast, our work proposes to learn a simple projection-based network that efficiently encodes intermediate network representations (i.e., hidden units) and operations involved, rather than the weights. We also introduce a new training paradigm for on-device models where the simple network is coupled and jointly trained to mimic an existing deep network that is flexible and can be customized by architecture or task. As we show in Section~\ref{sec:objfn}, the specifics of the training process can also include a choice to optimize towards soft targets as in model distillation approaches~\cite{distill2014}.

Dropouts~\cite{Srivastava2014} and other similar variants commonly used in practice for deep learning attempt to reduce parameters during neural network training by dropping unimportant neurons. However, they serve a different purpose, namely for better regularization.

\cite{WangLKC15} offers a survey of binary hashing literature that is relevant to the projection functions used in our work. The coupled network training architecture proposed in this paper (described in Section~\ref{sec:objfn}) also resembles, conceptually at a high level, generative adversarial networks (GANs)~\cite{goodfellow2014} which are used in unsupervised learning settings to reconstruct or synthesize data such photorealistic images.

\section{Neural Projection Networks}
\label{sec:randnn}
In this section, we present Neural Projection Networks, a joint optimization framework for training neural networks with reduced model sizes. We first introduce the objective function using a coupled full+projection network architecture and then describe the projection mechanism used in our work, namely locality sensitive hashing (LSH) and how it is applied here.

\subsection{ProjectionNets}
\label{sec:objfn}

Neural networks are a class of non-linear models that learn a mapping from inputs $\vec{x}_i$ to outputs $y_i$, where $\vec{x}_i$ represents an input feature vector or sequence (in the case of recursive neural networks) and $y_i$ is an output category for classification tasks or a predicted sequence. Typically, these networks consist of multiple layers of hidden units or neurons with connections between a pair of layers. For example, in a fully-connected feed-forward neural network, the number of weighted connections or network parameters that are trained is $O(n^2)$, where $n$ is the number of hidden units per layer.
\begin{figure*}[ht!]
\centering
\includegraphics[width=0.65\textwidth]{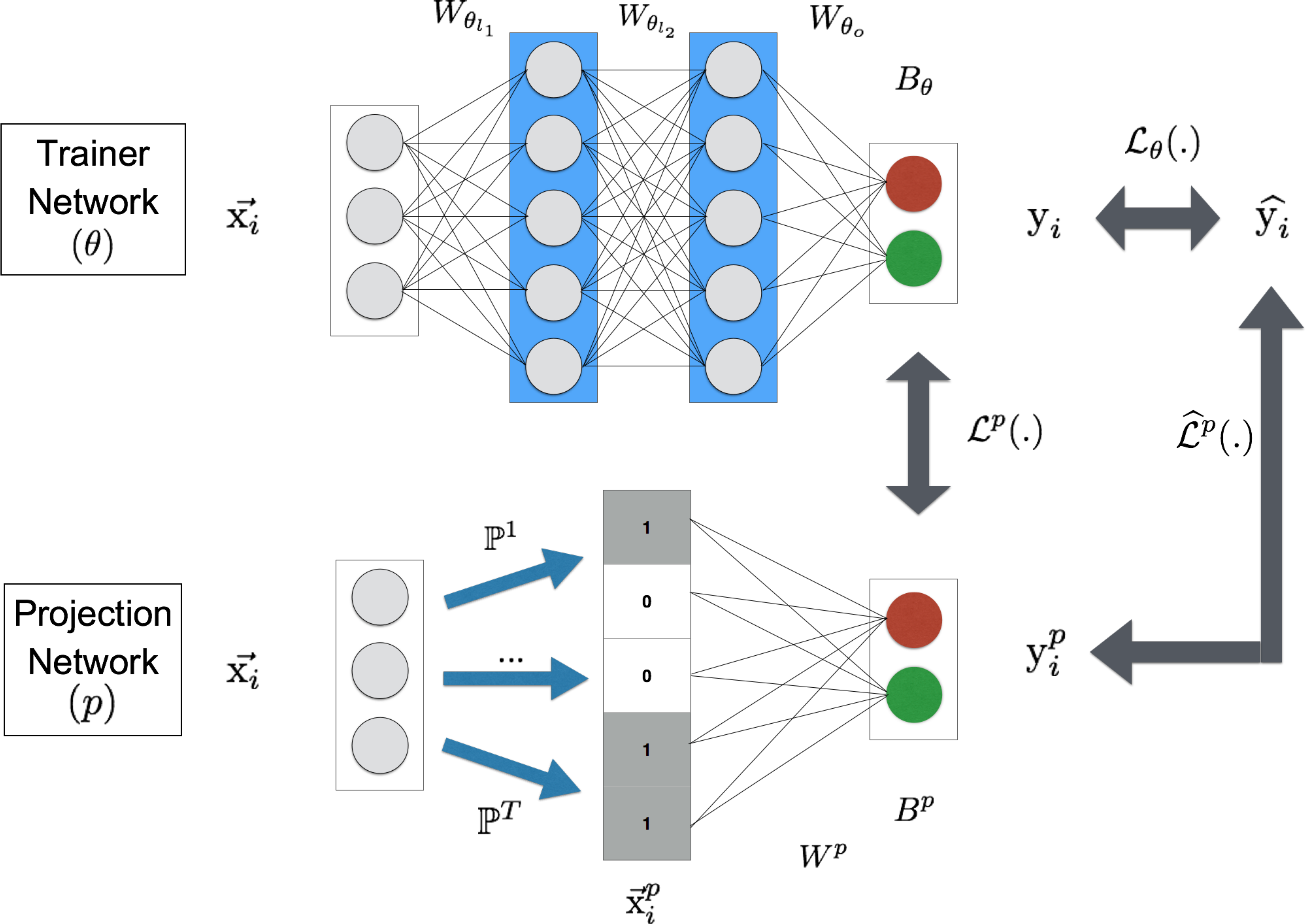}
\caption{Illustration of a Neural Projection Network trained using feed-forward NN.\hspace{\textwidth} {\it Notation:} $\vec{\textrm{x}}_i$ represents the input feature vector, $\widehat{\textrm{y}}_i$ the ground-truth, $y_i$ the prediction from the full network and $y^p_i$ the prediction from projection network. $\mathbb{P}^1$...$\mathbb{P}^T$ denote the $T$ projection functions that transform the input $\vec{\textrm{x}}_i$ into $d$-bit vectors, one per function. $W_{\theta}$, $B_\theta$ and $W^p$, $B^p$ represent the weights/bias parameters for the trainer network and projection network, respectively.\hspace{\textwidth} The training objective optimizes a combination of NN loss $\mathcal{L}_\theta(.)$ and projection loss $\mathcal{L}^p(.)$ that biases the projection network to mimic and learn from the full trainer network. The objective also incorporates a labeled loss $\widehat{\mathcal{L}}^p$ for the projection network.\label{fig:pnet}}
\end{figure*}

We propose a new objective and joint optimization framework for training compact on-device models for inference. The architecture uses a {\it trainer} network coupled with a {\it projection} network and trains them jointly. Figure~\ref{fig:pnet} illustrates the Neural Projection Network architecture using a feed-forward NN for the trainer network. The coupled networks are jointly trained to optimize a combined loss function:
\vspace{-0.5em}
\begin{align}
\mathcal{L}({\theta, p}) &= \lambda_1 \cdot \mathcal{L}_\theta(.) + \lambda_2 \cdot \mathcal{L}^p(.) + \lambda_3 \cdot \widehat{\mathcal{L}}^p(.)
\label{eqn:combined_projection_net}
\end{align}
where $\mathcal{L}_{\theta}(.)$, $\mathcal{L}^p(.)$ and $\widehat{\mathcal{L}}^p(.)$ are the  loss functions corresponding to the two networks as defined below. 
\begin{align}
\mathcal{L}_\theta(.)  &= \sum_{i \in N} \mathcal{D}(h_\theta(\vec{x}_i), \widehat{y}_i) \nonumber \\ 
\mathcal{L}^p(.)  &= \sum_{i \in N} \mathcal{D}(h^p(\vec{x}_i), h_\theta(\vec{x}_i)) \nonumber \\ 
\widehat{\mathcal{L}}^p(.)  &= \sum_{i \in N} \mathcal{D}(h^p(\vec{x}_i), \widehat{y}_i) 
\label{eqn:ind_projection_net}
\end{align}
\vspace{-1em}

$N$ indicates the number of training instances in the dataset, $\vec{x}_i$ represents the input feature vector in a feed-forward network or sequence input in an RNN, and $\widehat{y}_i$ refers to the ground-truth output classes used for network training. $h_\theta(\vec{x}_i)$ represents a parameterized representation of the hidden units in the {\it trainer} network that transforms $\vec{x}_i$ to an output prediction $y_i$. Similarly, $h^p(\vec{x}_i)$ represents the {\it projection} network parameters that transforms the input to corresponding predictions $y^p_i$. We apply softmax activation at the last layer of both networks to compute the predictions $y_i$ and $y^p_i$.

$\mathcal{D}$ denotes a distance function that measures the prediction error used in the loss functions. This is decomposed into three parts---trainer prediction error, projection simulation error and projection prediction error. Reducing the first leads to a better trainer network and decreasing the latter in turn learns a better projection network that is simpler but with approximately equivalent predictive capacity. In practice, we use cross-entropy for $\mathcal{D}(.)$ in all our experiments. For the {\it projection} $\mathcal{L}^p$ in Equation~\ref{eqn:ind_projection_net}, we follow a distillation approach~\cite{distill2014} to optimize $\mathcal{D}(.)$ since it has been shown to yield better generalization ability than a model trained on just the labels $\widehat{y}_i$.
$\lambda_1$, $\lambda_2$ and $\lambda_3$ are hyperparameters that affect the trade-off between these different types of errors. These are tuned on a small heldout development set and in our experiments, we set them to $\lambda_1=1.0, \lambda_2=0.1, \lambda_3=1.0$.

\noindent {\bf Trainer Network ($\theta$).} The {\it trainer} model is a full neural network (feed-forward, RNN or CNN) whose choice is flexible and depends on the task. Figure~\ref{fig:pnet} shows a trainer using feed-forward network but this can be swapped with LSTM RNNs (as we show later in Section~\ref{sec:semclass}) or other deep neural networks. For the network shown in the figure, the activations for $h_\theta(.)$ in layer $l_{k+1}$ is computed as follows:
\vspace{-1em}
\begin{align}
A_{\theta_{l_{k+1}}} &= \sigma(W_{\theta_{l_{k+1}}} \cdot A_{\theta_{l_{k}}} + B_{\theta_{l_{k+1}}})
\label{eqn:trainer_net}
\end{align}
where $\sigma$ is the ReLU activation function~\cite{NairH10} applied at each layer except the last and $A$ indicates the computed activation values for hidden units.
 \vspace{-0.5em}
 
The number of weights/bias parameters $W_\theta, B_\theta$ in this network can be arbitrarily large since this will only be used during the training stage which can be effectively done using high-performance distribtuted computing with CPUs or GPUs.

\noindent {\bf Projection Network ($p$).} The {\it projection} model is a simple network that encodes a set of efficient-to-compute operations which will be performed directly on device for inference. The model itself defines a set of efficient ``projection'' functions $\mathbb{P}(\vec{x}_i)$ that project each input instance $\vec{x}_i$ to a different space $\Omega_{\mathbb{P}}$ and then performs learning in this space to map it to corresponding outputs $y^p_i$. We use a simplified {\it projection} network with few operations as illustrated in Figure~\ref{fig:pnet}. The inputs $\vec{x}_i$ are transformed using a series of $T$ projection functions $\mathbb{P}^1, ..., \mathbb{P}^T$, which is then followed by a single layer of activations.
\vspace{-1em}
\begin{align}
\vec{x}^p_i &= \mathbb{P}^1(\vec{x}_i), ..., \mathbb{P}^T(\vec{x}_i)
\label{eqn:proj_fn}
\end{align}
\vspace{-3em}

\begin{align}
y^p_i &= \textrm{softmax}(W^p \cdot \vec{x}^p_i + B^p)
\label{eqn:proj_net}
\end{align}
The projection transformations use pre-computed parameterized functions, i.e., they are not trained during the learning process, and their outputs are concatenated to form the hidden units for subsequent operations. During training, the simpler {\it projection} network learns to choose and apply specific projection operations $\mathbb{P}^j$ (via activations) that are more predictive for a given task. It is possible to stack additional layers connected to the bit-layer in this network to achieve non-linear combinations of projections.

The {\it projection} model is jointly trained with the {\it trainer} and learns to mimic predictions made by the full trainer network which has far more parameters and hence more predictive capacity. Once learning is completed, the transform functions $\mathbb{P}(.)$ and corresponding trained weights $W^p$, $B^p$ from the projection network are extracted to create a lightweight model that is pushed to device. At inference time, the lightweight model and corresponding operations is then applied to a given input $\vec{x}_i$ to generate predictions $y^p_i$.

The choice of the type of projection matrix $\mathbb{P}$ as well as representation of the projected space $\Omega_{\mathbb{P}}$ in our setup has a direct effect on the computation cost and model size. We propose to leverage an efficient randomized projection method using a modified version of locality sensitive hashing (LSH) to define $\mathbb{P}(.)$. In conjunction, we use a bit representation $\mathbf{1}^d$ for $\Omega_{\mathbb{P}}$, i.e., the network's hidden units themselves are represented using projected bit vectors. This yields a drastically lower memory footprint compared to the full network both in terms of number and size of parameters. We highlight a few key properties of this approach below:
\begin{itemize}
\vspace{-0.5em}
\item There is no requirement for committing to a preset vocabulary or feature space unlike typical machine learning methods which resort to smaller vocabulary sizes as a scaling mechanism. For example, LSTM RNN models typically apply pruning and use smaller, fixed-size vocabularies in the input encoding step to reduce model complexity.
\item The proposed learning method scales efficiently to large data sizes and high dimensional spaces. This is especially useful for natural language applications involving sparse high dimensional feature spaces. For dense feature spaces (e.g., image pixels), existing operations like fully-connected layers (or even convolutions) can be efficiently approximated for prediction without relying on a large number of parameters. Such operations can also be applied in conjunction with the projection functions to yield more complex {\it projection} networks while constraining the memory requirements.
\item Computation of $\mathbb{P}(x_i)$ is independent of the training data size.
\item We ensure that $\mathbb{P}(.)$ is efficient to compute on-the-fly for inference on device. 
\end{itemize}
\vspace{-0.5em}
Next, we describe the projection method and associated operations in more detail.
\vspace{-0.5em}

\subsection{Locality Sensitive Projection Network}
\label{sec:lsh}

The {\it projection} network described earlier relies on a set of transformation functions $\mathbb{P}$ that project the input $\vec{x}_i$ into hidden unit representations $\Omega_{\mathbb{P}}$. The projection operations outlined in Equation~\ref{eqn:proj_fn} can be performed using different types of functions. One possibility is to use feature embedding matrices pre-trained using word2vec~\cite{word2vec} or similar techniques and model $\mathbb{P}$ as a embedding lookup for features in $\vec{x}_i$ followed by an aggregation operation such as vector averaging. However, this requires storing the embedding matrices which incurs additional memory complexity.  
  
Instead, we employ an efficient randomized projection method for this step. We use locality sensitive hashing (LSH)~\cite{Charikar2002} to model the underlying projection operations. LSH is typically used as a dimensionality reduction technique for applications like clustering~\cite{Manning2008}. Our motivation for using LSH within Projection Nets is that it allows us to project similar inputs $\vec{x}_i$ or intermediate network layers into hidden unit vectors that are nearby in metric space. This allows us to transform the inputs and learn an efficient and compact network representation that is only dependent on the inherent dimensionality (i.e., observed features) of the data rather than the number of instances or the dimensionality of the actual data vector (i.e., overall feature or vocabulary size). We achieve this with binary hash functions~\cite{Charikar2002} for $\mathbb{P}$.
\vspace{-0.5em}
\begin{theorem}
For $\vec{x}_i, \vec{x}_j \in \mathbb{R}^n$ and vectors $\mathbb{P}_k$ drawn from a spherically symmetric distribution on $\mathbb{R}^n$ the relation between signs of inner products and the angle $\measuredangle(\vec{x}_i,\vec{x}_j)$ between vectors can be expressed as follows:
\end{theorem}
\vspace{-1em}
\begin{align}
\measuredangle(\vec{x}_i, \vec{x}_j) &= \pi \hspace{4pt} \textrm{Pr} \{sgn [\langle\vec{x}_i, \mathbb{P}_k\rangle ] \neq sgn[ \langle\vec{x}_j, \mathbb{P}_k\rangle ] \}
\label{eqn:lsh_eqn}
\end{align}
This property holds from simple geometry~\cite{Charikar2002}, i.e., whenever a row vector from the projection matrix $\mathbb{P}$ falls inside the angle between the unit vectors in the directions of $\vec{x}_i$ and $\vec{x}_j$, they will result in opposite signs. Any projection vector that is orthogonal to the plane containing $\vec{x}_i \vec{x}_j$ will not have an effect. Since inner products can be used to determine parameter representations that are nearby, $\langle \vec{x}_i, \vec{x}_j \rangle = ||\vec{x}_i|| \cdot ||\vec{x}_j|| \cdot cos\measuredangle(\vec{x}_i, \vec{x}_j)$, therefore we can model and store the network hidden activation unit vectors in an efficient manner by using the signature of a vector in terms of its signs.

\noindent {\bf Computing Projections.} Following the above property, we use binary hashing repeatedly and apply the projection vectors in $\mathbb{P}$ to transform the input $\vec{x}_i$ to a binary hash representation denoted by $\mathbb{P}_k(\vec{x}_i) \in \{0, 1\}^d$, where $[\mathbb{P}_k(\vec{x}_i)] := sgn[\langle \vec{x}_i, \mathbb{P}_k \rangle]$. This results in a $d$-bit vector representation, one bit corresponding to each projection row $\mathbb{P}_{k= 1...d}$. 

The projection matrix $\mathbb{P}$ is fixed prior to training and inference. Note that we never need to explicitly store the random projection vector $\mathbb{P}_k$ since we can compute them on the fly using hash functions rather than invoking a random number generator. In addition, this also permits us to perform projection operations that are linear in the {\it observed} feature size rather than the {\it overall} feature size which can be prohibitively large for high-dimensional data, thereby saving both memory and computation cost. The binary representation is signficant since this results in a significantly compact representation for the {\it projection} network parameters that in turn reduces the model size considerably compared to the {\it trainer} network. 
  
Note that other techniques like quantization or weight sharing~\cite{CourbariauxBD14} can be stacked on top of this method to provide small further gains in terms of memory reduction.

\noindent {\bf Projection Parameters.} In practice, we employ $T$ different projection functions $\mathbb{P}^{j=1...T}$ as shown in Figure~\ref{fig:pnet}, each resulting in $d$-bit vector that is concatenated to form the projected activation units $\vec{x}^p_i$ in Equation~\ref{eqn:proj_fn}. $T$ and $d$ vary depending on the {\it projection} network parameter configuration specified for $\mathbb{P}$ and can be tuned to trade-off between prediction quality and model size.  
\vspace{-1em}
\subsection{Training and Inference}
\label{sec:train_infer}
\vspace{-1em}
We use the compact bit units to represent the {\it projection} network as described earlier. During training, this network learns to move the gradients for points that are nearby to each other in the projected bit space $\Omega_{\mathbb{P}}$ in the same direction. The direction and magnitude of the gradient is determined by the {\it trainer} network which has access to a larger set of parameters and more complex architecture. The two networks are trained jointly using backpropagation. Despite the joint optimization objective, training can progress efficiently with stochastic gradient descent with distributed computing on high-performance CPUs or GPUs.

Once trained, the two networks are de-coupled and serve different purposes. The {\it trainer} model can be deployed anywhere a standard neural network is used. The simpler {\it projection} network model weights along with transform functions $\mathbb{P}(.)$ are extracted to create a lightweight model that is pushed to device. This model is used directly ``on'' device at inference time by applying the same operations in Equations~\ref{eqn:proj_fn},~\ref{eqn:proj_net} (details described in Sections~\ref{sec:objfn},~\ref{sec:lsh}) to a new input $\vec{x}_i$ and generate predictions $y^p_i$. 

\noindent {\bf Complexity.} The overall complexity for inference is $O(n \cdot T \cdot d)$, where $n$ is the observed feature size (*not* overall vocabulary size) which is linear in input size, $d$ is the number of LSH bits specified for each projection vector $\mathbb{P}_k$, and $T$ is the number of projection functions used in $\mathbb{P}$. The model size (in terms of number of parameters) and memory storage required for the {\it projector} inference model in this setup is $O(T \cdot d)$.

\noindent {\bf Other Extensions.} It is possible to extend the proposed framework to handle more than one type of {\it trainer} or {\it projection} network and even simultaneously train several models at multiple resolutions using this architecture We leave this for future work.

As an alternative to the bit vector representation $\Omega_{\mathbb{P}}$, the projection matrix $\mathbb{P}$ can instead be used to generate a sparse representation of hidden units in the {\it projection} network. Each $d$-bit block can be encoded as an integer instead of a bit vector. This results in a larger parameter space overall $O(T \cdot 2^d)$ but can still be beneficial to applications where the actual number of learned parameters is tiny and inference can be performed via efficient sparse lookup operations.

\section{Experiments}
\label{sec:exp_rnn}
In this section we demonstrate the effectiveness of the proposed approach with several experiments on different benchmark datasets and classification tasks involving visual recognition and language understanding. The experiments are done using a TensorFlow implementation~\cite{tensorflow2015-whitepaper}.

\noindent {\bf Baselines and Method.} We compare ProjectionNets at different model sizes with full-sized deep neural network couterparts for each task. The deep network architecture varies depending on the task type--feed-forward network are employed for visual tasks whereas recursive neural networks (LSTMs) are used for language understanding task. In each case, the full neural network model used as baseline is also employed as {\it trainer} network to learn the small-sized ProjectionNets.

\noindent {\bf Evaluation.} For each task, we compute the performance of each model in terms of precision at different ranks K, i.e., accuracy \% within the top K predicted output classes. Models were trained using multiple runs, each experiment was run for a fixed number of (400k) time steps with a batch size of 200 for the visual tasks and 100 for the text classification task. The observed variance across runs wrt accuracy was small, around $\pm0.1\%$.

We also compute the {\it Compression Ratio} =$\frac{\textrm{\# parameters in Baseline deep network}}{\textrm{\# parameters in proposed model}}$ achieved by varying-sized models compared to the baseline deep network model.

The model size ratios reported here are based on number of free parameters and not wrt actual model size stored on disk. Note that this is slightly unfair to the ProjectionNet models and overlooks some of its advantage since the internal bit representations are far more efficient to store compactly on disk than standard neural network representations.
\subsection{MNIST Handwritten Digit Classification}
\label{sec:mnist_task}
We compare the performance of different approaches using the original MNIST handwritten digit dataset~\cite{mnistdata}. The dataset contains 60k instances for training and 10k instances for testing. We hold out 5k instances from the training split as dev set for tuning system parameters.

For the baseline and {\it trainer} network, we use a feed-forward NN architecture (3 layers, 1000 hidden units per layer) with L2-regularization.
{
\setlength{\textfloatsep}{-0.3em}
Table~\ref{tab:mnist} shows results of the baseline and comparison to ProjectionNet models with varying sizes ($T, d$). The results demonstrate that a tiny ProjectionNet with a remarkably high compression ratio of ~388x is able to achieve a high accuracy of 92.3\% compared to 98.9\% for the baseline that is significantly larger in memory footprint. Moreover, ProjectionNet models are able to achieve even further reduction in model size (upto 2000x-3500x) while yielding around 70-80\% precision for top-1 and 90-94\% precision among top-3 predictions.

\noindent {\bf Going deeper with projections:} Furthermore, going deeper with the projection network architecture (i.e., adding more layers) improves prediction performance even though it adds more parameters, overall size still remains significantly small compared to the baseline model. For example, a single-layer ProjectionNet with $T=60, d=10$ produces an accuracy of 91.1\% whereas a 2-layer deep ProjectionNet with the same projection layer followed by a fully connected layer (128 hidden units) improves the accuracy considerably to 96.3\%, yielding a gain of +5.2\%. A stacked ProjectionNet with slightly more parameters further improves the accuracy to 97.1\%, very close to the baseline performance at a 13x reduction in size.

\noindent {\bf Different training objectives:} The training objective can be varied by removing specific loss components from Equation~\ref{eqn:combined_projection_net}. We observe that using the joint architecture helps significantly and results in the best performance. The ProjectionNet $[T=60,d=12]$ from Table~\ref{tab:mnist} achieves 92.3\%. The same ProjectionNet trained {\it without} the full joint objective does worse, $\sim$91\% when trained using $\widehat{\mathcal{L}}^p$ alone or using only $L_\theta$ + $L^p$. We also trained smaller baseline neural networks with fewer layers and parameters for comparison and observed that ProjectionNets achieve far higher compression ratios at similar performance levels. The same trend follows when comparing against other simpler regularized linear baseline methods which perform far worse and produce accuracies comparable or worse than projection networks trained in isolation without the joint architecture. We notice that on more complex problems involving large output space, pre-training the trainer network and then performing joint training helps the projection net converge faster to a good performance.

\begin{table}[!ht]
\centering
\caption{Results for MNIST Classification using Neural Projection Nets and baselines.\label{tab:mnist}}
\scalebox{0.7}{
\begin{tabular}{|c||c||c|c|c|}
\hline
Model & Compression Ratio & Precision@1 & Precision@3 & Precision@5 \\ \hline \hline
NN  & 1 & 98.9 & 99.9 & 100.0 \\ \hline
ProjectionNet & & & &  \\  
 $[ T=8,d=10 ]$  & 3453 & 70.6 & 89.6 & 96.5  \\  
 $[ T=10,d=12 ]$  & 2312 & 76.9 & 93.9 & 98.1  \\  
 $[ T=60,d=10 ]$  & 466 & 91.1 & 98.7 & 99.6  \\  
 $[ T=60,d=12 ]$  &  388 & 92.3 &  99.0 &  99.6 \\ 
 $[ T=60,d=10 ]$ + FullyConnected [128 units] & 36 & 96.3 &  99.5 &  99.9 \\  
 $[ T=60,d=12 ]$ + FullyConnected [256 units] & 15 & 96.9 &  99.7  & 99.9 \\  
 $[ T=70,d=12 ]$ + FullyConnected [256 units] & 13 & 97.1 &  99.8  & 99.9 \\ \hline  
\end{tabular}}
\end{table}
}

\subsection{CIFAR-100 Image Classification}
\label{sec:cifarclass}
Next we evaluate the proposed approach on the CIFAR-100~\cite{cifardata} image classification task. The CIFAR-100 training dataset consists of 50k 32x32 color images in 100 classes, each with 500 images. Each image is tagged with a fine-grained class that it belongs to (e.g., {\it dolphin, seal, tulips, orchids}). The test dataset comprises of a separate 10k images, with 100 images per class.

This task is much harder than MNIST digit classification. Our goal for this setup is not an attempt to build a new image recognition system and outperform existing state-of-the-art which would require a lot of complex task-specific architecture customizations and convolution operations. Instead, we aim to train a standard deep neural network for this task and contrast it with various ProjectionNets to determine how much compression can be achieved and at what quality trade-off compared to the baseline.

We use a feed-forward NN architecture (3 layers, 1000 hidden units per layer) for the baseline and {\it trainer} network. Table~\ref{tab:cifar} shows results comaprisons between ProjectionNets and the baseline. As mentioned, this task is more complex than MNIST hence precision numbers are lower. However, we observe that a ProjectionNet model [T=60, d=12] is able to achieve more than 70x model compression with 33\% relative reduction in precision@5 compared to the baseline. The model size can be further reduced, upto 430x [T=10,d=12], while incurring an overall 42\% relative reduction in precision@5. On the other hand, a deeper projection network significantly reduces this performance gap to just 24\% while achieving a 49x model compression rate.

\begin{table}[!ht]
\centering
\caption{Results for CIFAR-100 Image Classification using Neural Projection Nets and baselines.\label{tab:cifar}}
\scalebox{0.7}{
\begin{tabular}{|c||c|c|c|}
\hline
Model & Precision@1 & Precision@3 & Precision@5 \\ \hline \hline
NN  & 32.9 & 50.2 & 58.6 \\ \hline
ProjectionNet & & &  \\  
 $[ T=10,d=12 ]$ & 12.8 &  25.2 & 33.8   \\  
 $[ T=60,d=12 ]$  &  17.8 & 30.1 &  39.2 \\  
 $[T=60,d=12]$ + FullyConnected [128 units]  &  20.8 & 35.6 &  44.8 \\ \hline 
\end{tabular}}
\end{table}
\subsection{Semantic Intent Classification}
\label{sec:semclass}
We compare the performance of the neural projection approach for training RNN sequence models (LSTM) for a semantic intent classification task as described in the recent work on SmartReply \cite{smartreply2016} for automatically generating short email responses. One of the underlying tasks in SmartReply is to discover and map short response messages to semantic intent clusters.\footnote{For details regarding SmartReply and how the semantic intent clusters are generated, refer \cite{smartreply2016}.} We choose 20 intent classes and created a dataset comprised of 5,483 samples (3,832 for training, 560 for validation and 1,091 for testing). Each sample instance corresponds to a short response message text paired with a semantic intent category that was manually verified by human annotators. For example, {\it``That sounds awesome!''} and {\it``Sounds fabulous''} belong to the {\it sounds good} intent cluster. 

We use an RNN sequence model with multilayer LSTM architecture (2 layers, 100 dimensions) as the baseline  and {\it trainer} network. The LSTM model and its ProjectionNet variant are also compared against other baseline systems---{\it Random} baseline ranks the intent categories randomly and {\it Frequency} baseline ranks them in order of their frequency in the training corpus. We show in table \ref{tab:rnn} that a ProjectionNet trained using LSTM RNNs achieves 82.3\% precision@1, only 15\% relative drop compared to the baseline LSTM but with significant reduction in memory footprint and computation (compared to LSTM unrolling steps). At higher ranks, the ProjectionNets achieve very high accuracies (90+\%) close to the baseline LSTM performance.
\setlength\textfloatsep{0.5em}
\begin{table}[!ht]
\centering
\caption{Results for Semantic Intent Classification using Neural Projection Nets and baselines.\label{tab:rnn}}
\scalebox{0.7}{
\begin{tabular}{|c||c|c|c|}
\hline
Model & Precision@1 & Precision@3 & Precision@5 \\ \hline \hline
Random & 5.2 & 15.0 & 27.0 \\ \hline
Frequency & 9.2 & 27.5 & 43.4 \\ \hline
LSTM  & 96.8 & 99.2 & 99.8 \\ \hline
LSTM-ProjectionNet & & &   \\  
 $[ T=60,d=12 ]$  & 82.3 & 93.5 & 95.5 \\ \hline 
\end{tabular}}
\end{table}

\section{Computing Model Predictability in Neural Bits}
\label{sec:neural_bits}
We further study the notion of predictability of deep neural networks in the context of {\it neural projections}. More specifically, using the ProjectionNet framework described in Section~\ref{sec:objfn}, we formulate the questions: ``How many neural bits are required to solve a given task?'' ``How many bits are required to capture the predictive power of a given deep neural network?''

These motivate a series of studies on the different datasets. We show empirical results to answer these by computing the number of bits encoded in the {\it projection} network (Section~\ref{sec:objfn}) used in each task. Since the projection neural network architecture is represented in bits (i.e., output of projection operations result in bit vectors) and we only use a single layer of projections in our experiments, we can compute the total number of bits required to represent a specific ProjectionNet model. If we compare this to the accuracy achieved on a given task by applying the {\it projection} network alone for inference, this helps answers the first question. 

On the visual recognition task for MNIST (Section~\ref{sec:mnist_task}), 80-100 neural projection bits are enough to solve the task with 70-80\% accuracy, and increasing this to 720 bits achieves 92.3\% precision which is further improved by using a deeper projection network.  For the language task of semantic classification (Section~\ref{sec:semclass}) involving sequence input, 720 neural projection bits are required to achieve a top-1 accuracy of 82.3\%. 
{
To answer the second question, we use a given deep neural network (e.g, feed-forward NN) with specified configuration to model the full {\it trainer} network in the framework described in Section~\ref{sec:randnn}. We then use this to train a corresponding neural {\it projection} network with hidden bit representations. Finally, we compute the number of neural projection bits used in the second network that simulates the {\it trainer} and plot this value against the {\it predictive quality ratio}, which is defined as the ratio of accuracies achieved by separately performing inference with simple versus full network on a given task. Figure~\ref{fig:plot_nbit} shows this plot for the MNIST and CIFAR-100 tasks. The plot shows that the predictive power of a 3-layer feed-forward network with 3-5M parameters can be succinctly captured to a high degree (ratio=$\sim$0.8) with a simple 100-bit ProjectionNet for MNIST classification and just 720 bits are required to recover more than 90\% of the base deep network quality. On more complex image recognition tasks that involve larger output classes, a higher number of bits are required to represent a {\it trainer} deep network with the same architecture. For example, on the CIFAR-100 task, we observe that a 3-layer feed-forward network can be projected onto 720 neural bits at a predictive ratio of 0.5. However, we also notice a steep increase in predictive ratio moving from 120 to 720 neural bits. We expect a similar trend at even higher bit sizes and on more complex tasks.

\begin{figure*}[ht!]
\centering
\includegraphics[width=0.45\textwidth]{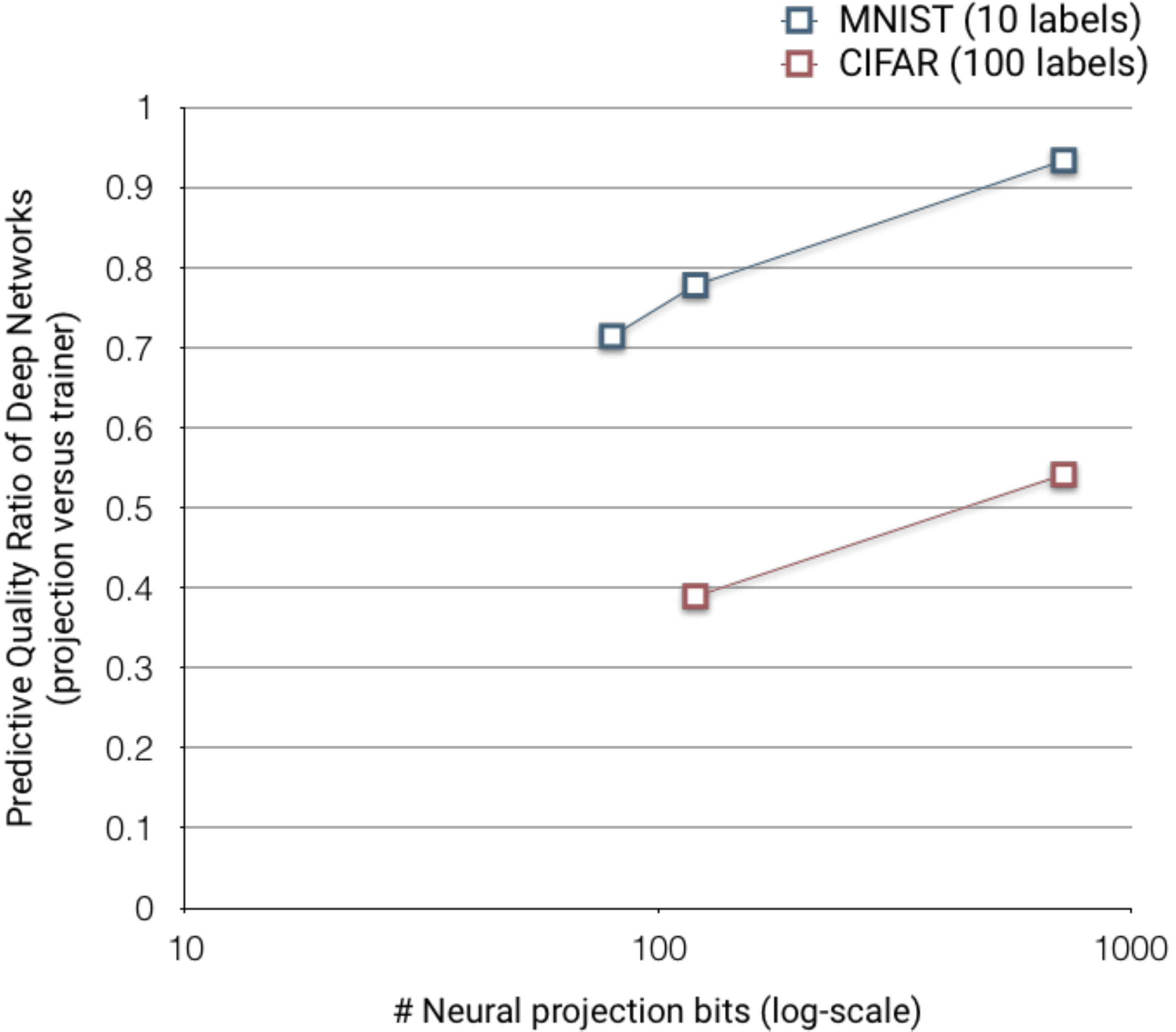}
\caption{Comparing predictive power of deep neural networks using neural projection bits on different visual classification tasks. \label{fig:plot_nbit}}
\end{figure*}
}
\begin{figure*}[ht!]
\centering
\includegraphics[width=0.55\textwidth]{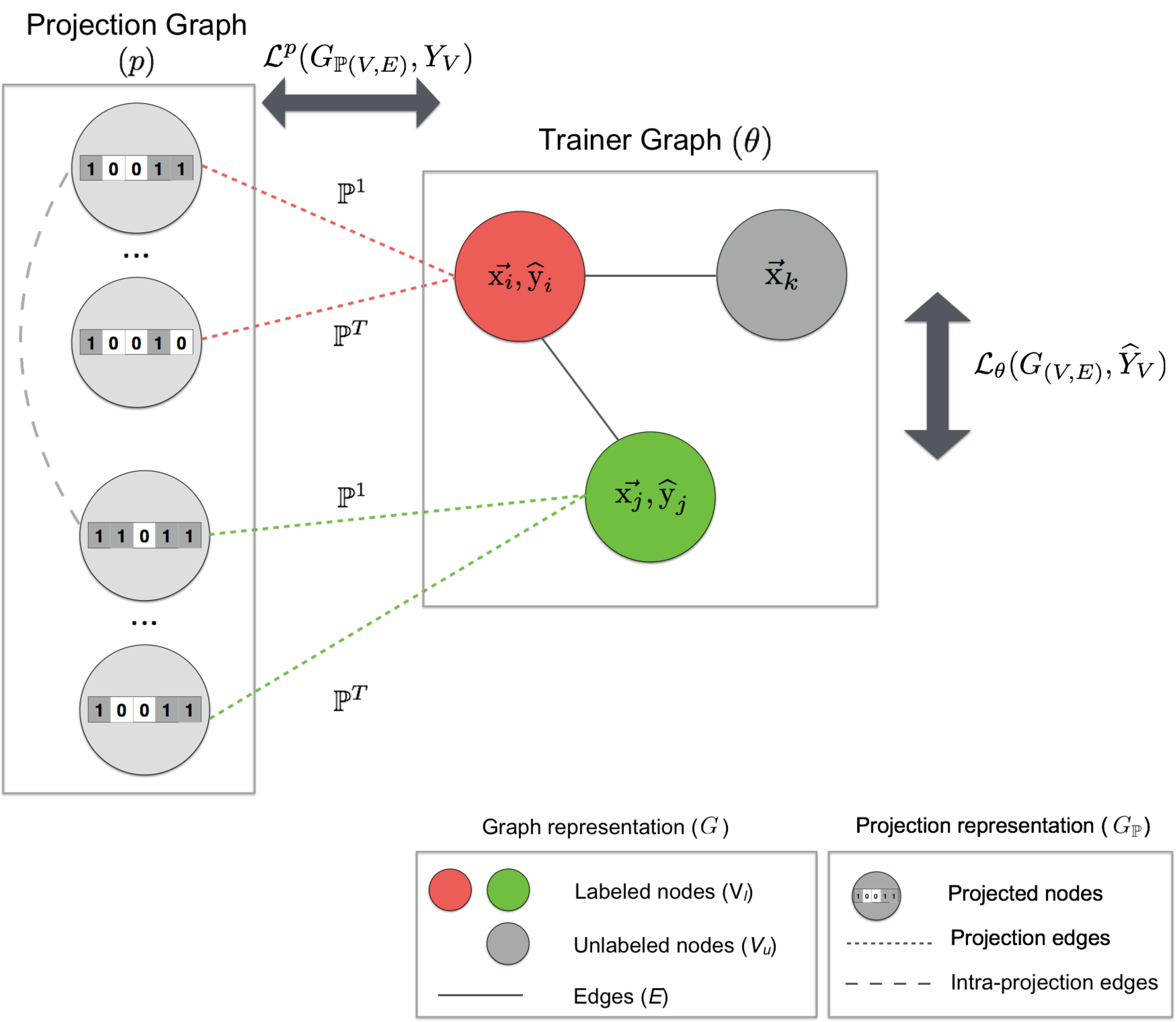}
\caption{Illustration of a Projection Graph trained using graph learning algorithms.\hspace{\textwidth}{\it Notation for trainer graph:} $\vec{\textrm{x}}_i$ represents an input node and $\vec{\textrm{x}}_j$, $\vec{\textrm{x}}_k$ represent its neighborhood $\mathcal{N}(\vec{\textrm{x}}_i)$ in the original graph $G = (V,E)$. $E$ refers to edges and $V = V_l \cup V_u$ the nodes in this graph, where $V_l$ indicates labeled nodes {\it (red, green)} and $V_u$ the unlabeled nodes {\it (grey)}. $\widehat{\textrm{y}}_i$ indicates the ground-truth corresponding to a labeled node $\vec{\textrm{x}}_i \in V_l$, with {\it red, green} colors indicating different output values for $\widehat{\textrm{y}}_i$. If $V=V_l$, the graph is trained in a supervised manner whereas $V_l \subset V; V_u \neq \varnothing$ yields a semi-supervised graph learning formulation.\hspace{\textwidth}{\it Notation for projection graph:} Each node $\vec{\textrm{x}}_i$ in the trainer graph $G$ is connected via an edge to a corresponding projection node in $G_\mathbb{P}$. The projection nodes are discrete representations of trainer node $\vec{\textrm{x}}_i$ obtained by applying the projection functions $\mathbb{P}^1...\mathbb{P}^T$ to the feature vector associated with the trainer node. In addition, $G_\mathbb{P}$ may also contain intra-projection edges computed using a similarity metric applied to the projection vector representations. For example, we can use Hamming distance $\mathcal{H}(.)$ to define a similarity metric $1-\frac{\mathcal{H}(.)}{d}$ between projection nodes represented as $d-$bit vectors. \hspace{\textwidth}The training objective optimizes a combination of graph loss $\mathcal{L}_\theta(.)$ and projection loss $\mathcal{L}^p(.)$ that biases the projection graph to mimic and learn from the full trainer graph. $\mathcal{L}_\theta(.)$ optimizes the trainer's predicted output $\textrm{y}_i$ against the ground-truth $\widehat{\textrm{y}}_i$ whereas $\mathcal{L}^p(.)$ optimizes predictions from the projection graph $\textrm{y}^p_i$ against the neigbhoring trainer predictions $\textrm{y}_i$.\label{fig:pgraph}}
\end{figure*}
\section{Discussion and Future Work}
\label{sec:conc}
We introduced a new Neural Projection approach to train lightweight neural network models for performing efficient inference on device at low computation and memory cost. We demonstrated the flexibility of this approach to variations in model sizes and deep network architectures. Experimental results on different visual and language classification tasks show the effectiveness of this method in achieving significant model size reductions and efficient inference while providing competitive performance. We also revisited the question of predictability of deep networks and study this in the context of neural bit projections.

A few possible future extensions to the framework are discussed at the end of Section~\ref{sec:randnn}. Going beyond deep learning, it is also possible to apply this framework to train lightweight models in other types of learning scenarios. For example, the training paradigm can be changed to a semi-supervised or unsupervised setting. The {\it trainer} model itself can be modified to incorporate structured loss functions defined over a graph or probabilistic graphical model instead of a deep neural network. Figure~\ref{fig:pgraph} illustrates an end-to-end {\it projection graph} approach to learning lightweight models using a graph optimized loss function which can be trained efficiently using large-scale distributed graph algorithms~\cite{ravi2016large} or even neural graph approaches~\cite{ngm2017,YangCS16}. The projection model training can also be further extended to scenarios involving distributed devices using complementary techniques~\cite{KonecnyMRR16}. We leave these explorations as future work.

\renewcommand\refname{\vskip -1cm}
\subsubsection*{References}
{\small
\bibliography{refs}{}
\bibliographystyle{ieeetr}
}
\end{document}